\documentclass{article}

\usepackage{PRIMEarxiv}

\usepackage[utf8]{inputenc} 
\usepackage[T1]{fontenc}    
\usepackage{hyperref}       
\usepackage{url}            
\usepackage{booktabs}       
\usepackage{amsfonts}       
\usepackage{nicefrac}       
\usepackage{microtype}      
\usepackage{lipsum}
\usepackage{fancyhdr}       
\usepackage{graphicx}       
\graphicspath{{media/}}     

\usepackage{bm}
\usepackage{adjustbox}

\title{PEDENet: Image Anomaly Localization via Patch Embedding and Density Estimation}

\author{
  Kaitai Zhang \\
  University of Southern California \\
  Los Angeles, USA\\
  \texttt{kaitaizh@usc.edu} \\
   \And
  Bin Wang \\
  National University of Singapore \\
  Singapore\\
  \texttt{bwang28c@gmail.com} \\
   \And
  C.-C. Jay Kuo \\
  University of Southern California \\
  Los Angeles, USA\\
  \texttt{cckuo@sipi.usc.edu} \\
}

\begin{document}
\maketitle

\begin{abstract}
A neural network targeting at unsupervised image anomaly localization,
called the PEDENet, is proposed in this work.  PEDENet contains a patch
embedding (PE) network, a density estimation (DE) network, and an
auxiliary network called the location prediction (LP) network. The PE
network takes local image patches as input and performs dimension
reduction to get low-dimensional patch embeddings via a deep encoder
structure. Being inspired by the Gaussian Mixture Model (GMM), the DE
network takes those patch embeddings, and then predicts the cluster
membership of an embedded patch.  The sum of membership probabilities is
used as a loss term to guide the learning process. The
LP network is a Multi-layer Perception (MLP), which takes embeddings
from two neighboring patches as input and predicts their relative
location. The performance of the proposed PEDENet is evaluated
extensively and benchmarked with that of state-of-the-art methods.
\end{abstract}


\section{Introduction}\label{sec:introduction}

Image anomaly detection is a binary classification problem that decides
whether an input image contains an anomaly or not. Image anomaly
localization is to further localize the anomalous region at the pixel
level.  Due to recent advances in deep learning and availability of new
datasets, recent research works are no longer limited to the image-level
anomaly detection result, but also show a significant interest in the
pixel-level localization of anomaly regions.  Image anomaly detection
and localization find real-world applications such as manufacturing
process monitoring\cite{scime2018anomaly}, medical image analysis
\cite{schlegl2017unsupervised, schlegl2019f}, and video surveillance
analysis \cite{saligrama2012video,zhou2019anomalynet}. 

Most well-studied localization solutions (e.g., semantic segmentation)
rely on heavy supervision, where a large number of pixel-level labels
and many labeled images are needed. However, in the context of image
anomaly detection and localization, a typical assumption is that only
normal (i.e., artifact-free) images are available in the training stage.
This is because anomalous samples are few in general.  Besides, they are
often hard to collect and expensive in labeling.  To this end, image
anomaly localization is usually done in an unsupervised manner, and
traditional supervised solutions are not applicable. 

Several methods that integrate local image features and anomaly
detection models for anomaly localization have been proposed recently,
e.g., \cite{rippel2020modeling, cohen2020sub,zhang2021anomalyhop}.  They first employ a
deep neural network to extract local image features and then apply the
anomaly detection technique to local regions. 
Generally speaking, stage-by-stage training is only sub-optimal.
However, when the stages are relatively independent, the loss could be
little. Actually, it is observed in Sec. \ref{sec:experiments} that this
approach offers excellent (and even the best) performance among various
benchmarking models \cite{rippel2020modeling,cohen2020sub}. This is
attributed to the adopt of a very large pretrained network.  On one
hand, the pre-trained network plays an important role in their superior
performance. On the other hand, it demands higher computational
complexity and memory requirement.  Thus, it is costly to deploy in
real-world applications.

To address these issues, we present a new neural network model, called
the PEDENet, for unsupervised image anomaly localization in this work.
PEDENet contains a patch embedding (PE) network, a density estimation
(DE) network and an auxiliary network called the location prediction
(LP) network.  The PE network utilizes a deep encoder to get a
low-dimensional embedding for local patches. 
Inspired by the Gaussian mixture model (GMM), the DE network models the
distribution of patch embeddings and computes the membership of a patch
belonging to certain modalities. It helps identify outlying artifact
patches. The sum of membership probabilities is used as a loss term to
guide the learning process.  The LP network is a Multi-layer Perception
(MLP), which takes patch embeddings as input and predicts the relative
location of corresponding patches. The performance of the proposed
PEDENet is evaluated and compared with state-of-the-art benchmarking
methods by extensive experiments. 

\begin{figure}[!t]
\includegraphics[width=0.9\linewidth]{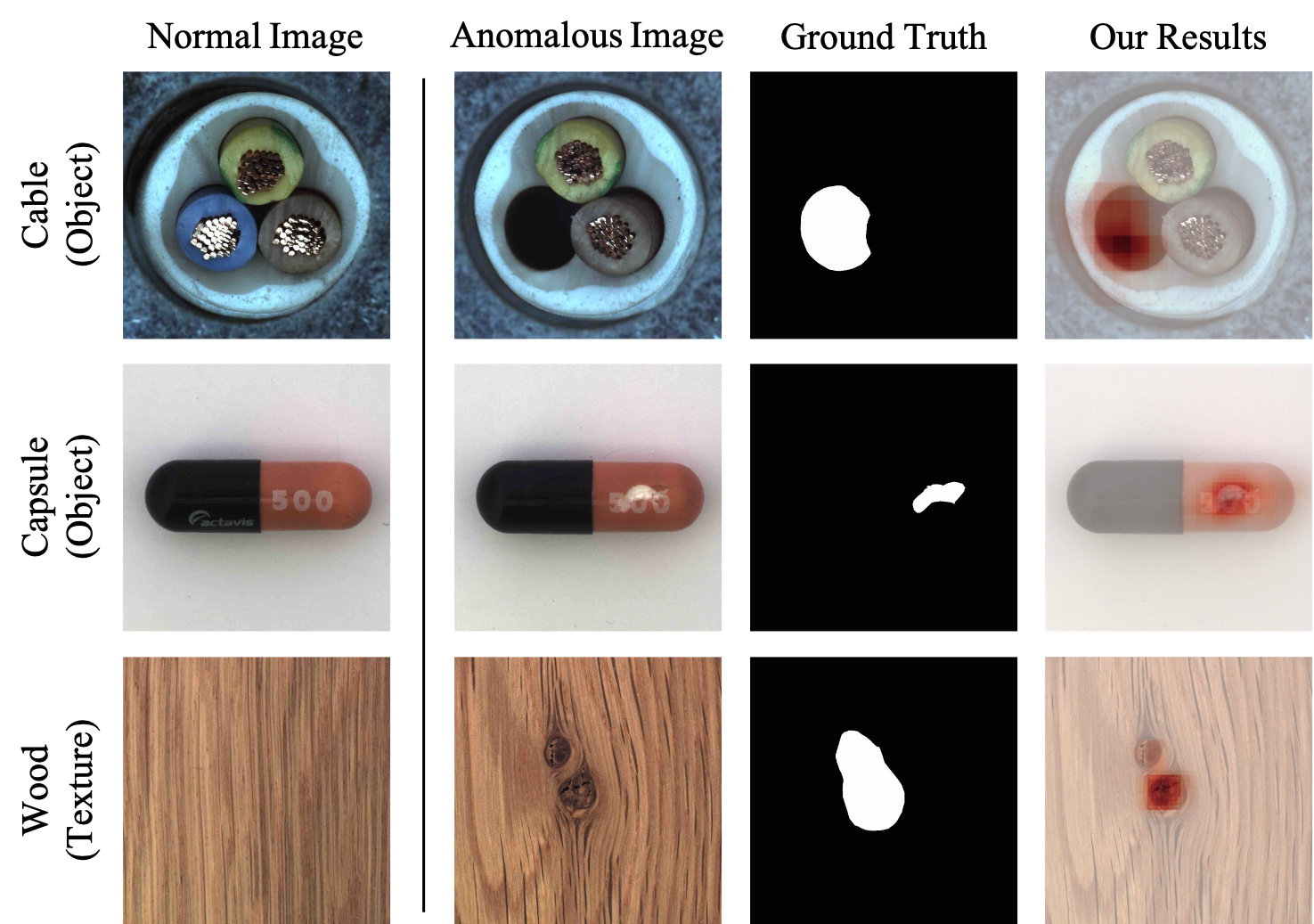}
\caption{Image anomaly localization examples (from left to right):
normal images, anomalous images, ground truth of the anomalous region
and the predicted anomalous region by the proposed PEDENet, where the
red region indicates the detected anomalous region. These examples are
taken from the MVTec AD dataset.}\label{fig:1}
\end{figure}

Our work has the following three main contributions.
\begin{itemize}
\item We propose PEDENet for unsupervised image anomaly localization. It
can be trained in an end-to-end manner with only normal images (i.e.
unsupervised learning). 
\item Being inspired by the GMM, the DE network models the distribution
of patch embeddings and computes the membership of a patch belonging to
certain modalities. It helps find outlying artifact patches.
\item Experiments show that PEDENet achieves state-of-the-art performance on
unsupervised anomaly localization for the MVTec AD dataset. 
\end{itemize}

The rest of this letter is organized as follows. Related previous work
is reviewed in Sec. \ref{sec:review}. The proposed PEDENet is presented
in Sec. \ref{sec:method}. Experimental results are shown in Sec.
\ref{sec:experiments}. Finally, concluding remarks are given in Sec.
\ref{sec:conclusion}.

\section{Related Work}\label{sec:review}
Some related previous work is reviewed in this section. For image
anomaly detection and localization, there are three commonly used
approaches: 1) reconstruction-based, 2) pretrained network-based and 3)
one-class classification-based approaches. They are reviewed in Secs.
\ref{subsec:reconstruction}, \ref{subsec:network} and
\ref{subsec:one-class}, respectively. Finally, work on non-image anomaly data
is briefly mentioned in Sec. \ref{subsec:non-image}.

\subsection{Reconstruction-based Approach}\label{subsec:reconstruction}

Since normal samples are only available in training, the
reconstruction-based approach utilizes the extraordinary capability of
neural networks to focus on normal samples' characteristics. 
Early work mainly uses the autoencoder and its variants
\cite{DBLP:conf/visapp/BergmannLFSS19, salehi2020puzzle,
liu2020towards, rudolph2021same}. Since they are trained with only
normal training data, it is unlikely for them to reconstruct abnormal
images in the testing stage. As a result, the pixel-wise difference
between the input abnormal image and its reconstructed image can
indicate the region of abnormality. 
However, this approach is problematic due to inaccurate reconstructions
or poorly calibrated likelihoods.  More recently, several methods
exploit factors on top of the reconstruction loss such as the
incorporation of an attention map \cite{venkataramanan2020attention} or
a student-teacher network to exploit intrinsic uncertainty in
reconstruction \cite{bergmann2020uninformed}.  A similar idea is to
adopt the inpainting model as an alternative \cite{li2020superpixel},
\cite{zavrtanik2021reconstruction}. It first removes part of an image
and then reconstructs the missing part based on the visible part. The
difference between the removed region and its inpainted result could
tell the abnormality level of that region.  Although they show promising
results, the resolution of anomaly maps is somehow coarse due to the
heavy computational burden. 

\subsection{Pretrained Network-based Approach}\label{subsec:network}

The performance of some computer vision algorithms can be improved by
transfer learning using discriminative embeddings from pretrained
networks. Along this line of thought, some models combine image features
obtained by pretrained networks with anomaly detection algorithms.  For
example, the nearest-neighbor algorithm is used in \cite{cohen2020sub}
to examine whether an image patch of a test image is similar to any
known normal image patches in the training set. The Gaussian
distribution is adopted by \cite{rippel2020modeling} 
to fit the distribution of local features that
are extracted from a pretrained network.  However, since the pretrained
networks are not specially optimized for the image anomaly detection
task, the resulting model usually has a large model size. 

\subsection{One-class Classification-based Approach}\label{subsec:one-class}

Another natural idea is to adopt the Support Vector Data Description
(SVDD) classifier. SVDD is a classic one-class classification algorithm
derived from the Support Vector Machine (SVM). It maps all normal
training data into a kernel space and seeks the smallest hypersphere
that encloses the data in the space. Anomalies are expected to be
located outside the learned hypersphere. Ruff {\em et al.}
\cite{ruff2018deep} first incorporated this idea in a deep neural
network for non-image data anomaly detection and then extended it to an
unsupervised setting \cite{ruff2019deep}. They used a neural network
to mimic the kernel function and trained it with the radius of the
hypersphere. This modification allows an encoder to learn a
data-dependent transformation, thus enhancing detection performance on
high-dimensional and structured data.  Later, Liznerski {\em et al.}
\cite{liznerski2020explainable} generalized it to image anomaly
detection by applying a SVDD-inspired pseudo-Huber loss to the output
matrix of a Fully Convolutional Network (FCN). It offers further
improvements in a semi-supervised setting. 

\begin{figure*}[t]
\centering
\includegraphics[width=1.0\linewidth]{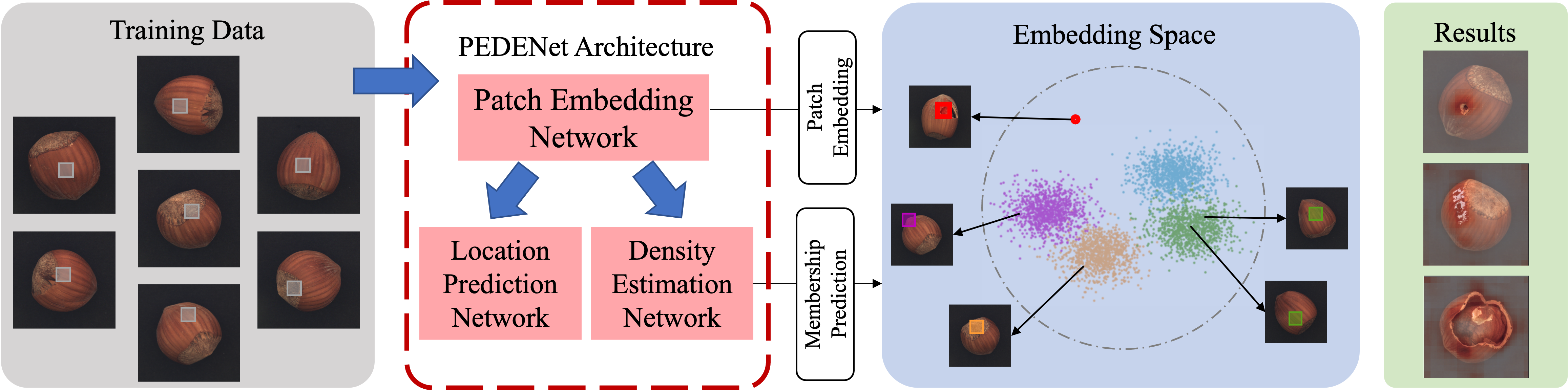}
\caption{An overview of the proposed PEDENet. Image are first divided
into patches and, then, fed into the Patch Embedding (PE) network for
patch embeddings. The Density Estimation (DE) network conducts clustering 
in the embedding space. After training, normal patches are clustered, and
outliers could be treated as abnormal patches in the inference stage. Three
anomaly localization results from thee Hazelnut class are shown as examples.}\label{fig:2a}
\end{figure*}

\subsection{Non-image Data}\label{subsec:non-image}

For anomaly detection on non-image data, a Deep Autoencoding Gaussian
Mixture Model(DAGMM) was proposed in \cite{zong2018deep}. It combines
dimensionality reduction and density estimation for unsupervised anomaly
detection. Our proposed PEDENet is different from DAGMM in three
aspects. First, PEDENet is designed for image anomaly localization while
DAGMM targets at identifying an anomalous entity. Second, DAGMM employs
an autoencoder (AE) to reconstruct the input data and concatenates
latent features with reconstruction errors for density estimation.
PEDENet adopts a PE network to learn local features and then applies the
DE network to patch embeddings. Third, DAGMM demonstrates its performance
on non-image data whose dimension is relatively lower. For example, the
dimension of the latent space can be as low as one. In contrast, the
dimension of patch embedding features is significantly higher in
PEDENet. 


\section{Proposed Method}\label{sec:method}

\subsection{PEDENet}\label{subsec:PEDENet}

\begin{figure}[!t]
\centering
\includegraphics[width=0.5\linewidth]{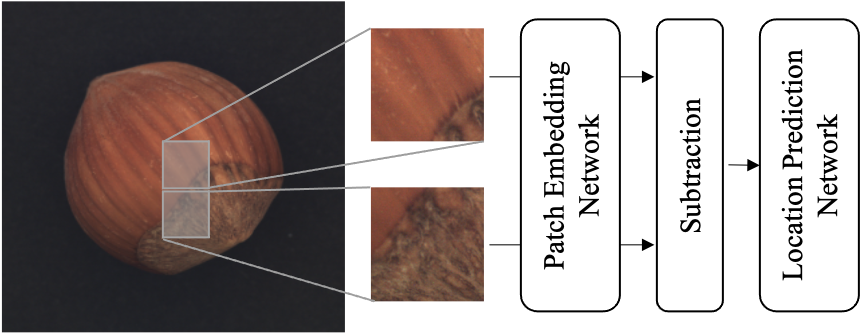}
\caption{The diagram of the location prediction (LP) network.}\label{fig:2b}
\end{figure}

An overview of the PEDENet is shown in Fig. \ref{fig:2a}, where the
Hazelnut class from the MVTec AD dataset is used as an illustrative input. 
PEDENet conducts class-specific training and testing.  It contains a
patch embedding (PE) network, a density estimation (DE) network, and an
auxiliary network called the location prediction (LP) network.  The PE
network takes an image patch as its input and performs dimension
reduction to get low-dimensional patch embeddings via a deep encoder
structure. The DE network takes a patch embedding as the input and
predicts its cluster membership under the framework of the Gaussian
Mixture Model (GMM).  These two networks are shown in Fig.
\ref{fig:2a}.  The LP network is a multi-layer perception (MLP).  It
takes a pair of patch embeddings as the input and predicts the relative
location of them as depicted in Fig. \ref{fig:2b}. 

{\bf PE Network.} With an image patch as the input, the PE network
outputs its low-dimensional embedding.  Inspired by \cite{yi2020patch},
we adopt a hierarchical encoder that embodies a larger encoder with
several smaller encoders. That is, we divide an input patch, $p$, into
$2 \times 2$ sub-patches and feed each of smaller sub-patches into a
smaller encoder.  The outputs of the smaller sub-patches are aggregated
based on their original positions. Then, they are encoded by a larger
encoder to get the embedding of patch $p$.  The above process can be
summarized as
\begin{equation}\label{eq0}
\bm{z} = \mbox{PEN}(\bm{p}; \bm{\theta}_{PEN}),
\end{equation}
where PEN denotes the PE network and $\bm{z} \in \mathbb{R}^Z $ is the 
low-dimensional embedding output of $p$, and $Z$ is the dimensionality 
of the embedding space. The PE network is implemented as a composite
mapping consisting of smaller encoders, a larger encoder and the
aggregation process with learnable parameters $\bm{\theta}_{PEN}$
as shown in the right-hand-side of Eq. (\ref{eq0}).


{\bf DE Network.} Given the low-dimensional embedding of an input patch,
the DE network predicts its cluster membership values with respect to
multiple Gaussian modalities under the Gaussian Mixture Model (GMM).  A
GMM assumes data samples are generated from a finite number of Gaussian
distributions with unknown parameters.  The Expectation-Maximization
(EM) algorithm can be used to optimize those parameters iteratively. In
the EM algorithm, the Expectation step computes the posterior
probability (i.e. the so-called cluster membership) of each sample while
the Maximization step updates the GMM parameters, including the mean
vector, the covariance matrix and the mixture coefficient of each
Gaussian component. 

Note that GMM may suffer from a sub-optimal solution. Furthermore, it is
challenging to combine it with a neural network to achieve end-to-end
learning. By following the idea in DAGMM \cite{zong2018deep}, we propose
the DE network to estimate the cluster membership for each patch
embedding, which corresponds to the expectation step of the EM algorithm:
\begin{equation}\label{eq1}
\bm{\gamma} =  \mbox{softmax}(\mbox{DEN}(\bm{z}; \bm{\theta}_{DEN})),
\end{equation}
where DEN indicates the DE network, $\bm{z}$ is the low-dimensional
embedding generated by the PE network, and $\bm{\theta}_{DEN}$ denotes
parameters of the DE network. After the softmax normalization,
$\bm{\gamma}$ is a vector in $\mathbb{R}^K$, where $K$ is a
hyper-parameter representing the number of Gaussian components in the
GMM. 

After getting the membership prediction $\bm{\gamma}$, we update GMM
parameters. This corresponds to the maximization step of the EM
algorithm.  Here, we use $\phi_k$, $\bm{\mu_k}$ and $\bm{\Sigma}_k$ to
represent the mixture coefficient, the mean vector, the covariance
matrix of the $k$-th component, $1 \leq k \leq K$. For a batch of $N$
patch embeddings, we have
\begin{eqnarray}
\phi_k & = & \sum^N_{i=1}\frac{\gamma_{ik}}{N} \in \mathbb{R}, \label{eq2} \\
\bm{\mu_k} & = & \frac{\sum^N_{i=1}\gamma_{ik}\bm{z_i}}
{\sum^N_{i=1}\gamma_{ik}} \in \mathbb{R}^Z, \label{eq3} \\
\bm{\sigma}_k & = &\frac{\sum^N_{i=1}\gamma_{ik}(\bm{z_i}-\bm{\mu_k})(\bm{z_i}
-\bm{\mu_k})^T}{\sum^N_{i=1}\gamma_{ik}}\in \mathbb{R}^{Z\times Z}. \label{eq4}
\end{eqnarray}
Then, we can express the probability of a patch embedding $\bm{z_i}$ in form of
\begin{equation}\label{eq5}
P(\bm{z_i}) = \sum^K_{k=1} \phi_k \frac{\exp(-\frac{1}{2}(\bm{z_i}
-\bm{\mu_k})\bm{\Sigma}_k^{-1}(\bm{z_i}-\bm{\mu_k})^{T})}
{\sqrt{2\pi |\bm{\Sigma}_k}|} \in [0, 1],
\end{equation}
where $|\cdot |$ means the determinant of a matrix. If a patch embedding,
$\bm{z_i}$, has a large probability, it means the corresponding patch
is likely to be a normal patch, and vice versa. 

{\bf LP Network.} Inspired by Patch-SVDD \cite{yi2020patch}, we
employee the LP network as an auxiliary network that predicts the
relative location of two neighboring patches in a self-supervised
manner. 
For an arbitrary input patch, $\bm{p}$, we sample another patch
$\bm{p}^{\prime}$ from one of its eight neighbors as shown in
Fig. \ref{fig:2b}. We feed patches $\bm{p}$ and $\bm{p}^{\prime}$ into
the PE network to get their patch embeddings, $\bm{z}$ and $\bm{z}^{\prime}$,
respectively. The relative location of patch $\bm{p}^{\prime}$ against 
patch $\bm{p}$ is encoded to a one-hot vector $\bm{l} \in \mathbb{R}^8$, 
used as the label in the training as:
\begin{equation}\label{eq6}
\bm{\hat{l}} = \mbox{LPN}(\bm{z}, \bm{z}^{\prime}; \bm{\theta}_{LPN}),
\end{equation}
where LPN denotes the LP network and $\bm{\theta}_{LPN}$ denotes its
network parameters. 

\subsection{Loss Function}\label{subsec:loss}

We propose the following loss function to train the PE, DE and LP 
networks jointly:
\begin{equation}\label{eq10}
\mathcal{L}= \lambda_1\mathcal{L}_{DEN} + \lambda_2\mathcal{L}_{LPN} + 
\lambda_3\mathcal{L}_{reg},
\end{equation}
where $\lambda_1$, $\lambda_2$, and $\lambda_3$ are parameters adjustable
by users to assign a different weight to each loss terms. The three
terms in the right of Eq. (\ref{eq10}) are elaborated below.

The first loss term is needed for the DE network. Since the total
probability, $P(\bm{z_i})$, models the likelihood of observing
$\bm{z_i}$ as a normal patch. By maximizing the average total
probability of input patches,
\begin{equation}\label{eq7}
\mathcal{L}_{DEN} = -\frac{1}{N}\log\sum^N_{i=1}P(\bm{z_i}),
\end{equation}
the DE network is trained to describe the distribution of patches
embedding with an implicit GMM while the PE network can be optimized
simultaneously.  The second loss term is the cross-entropy loss between
$l$ and $\hat{l}$ for the LP network as given in Eq. (\ref{eq7}):
\begin{equation}\label{eq8}
\mathcal{L}_{LPN} = -\sum^8_{i=1}l_i \cdot log(\hat{l_i}).
\end{equation}
Finally, we add a regularization term to prevent singularity in the 
GMM, which occurs when the determinant of any $\bm{\Sigma}_k$
degenerates to zero. We add the regularization loss term to
penalize small values of diagonal elements:
\begin{equation}\label{eq9}
\mathcal{L}_{reg} =\sum^K_{k=1}\sum^Z_{z=1}\frac{1}{\bm{\Sigma}_{kz}}.
\end{equation}

{\bf End-to-end Training.} With the total loss function given in Eq.
(\ref{eq10}), all three networks could be jointly optimized using the
back-propagation algorithm. The DE network and the LP network are two
parallel branches concatenated to the PE network. That is, the output of
the PE network, i.e. patch embedding, serves as the input to the DE and
the LP networks. The DE network takes a patch embedding directly. The LP
network takes the difference of embeddings of a pair of adjacent patches
and use their relative position as the training label.  In the training,
parameters of all three networks are updated simultaneously so as to
achieve end-to-end training for the whole PEDENet. 

\subsection{Anomaly Localization}\label{subsec:localization}

We obtain low-dimensional patch embeddings from the PE
network so as to localize anomalies.  Low-dimensional patch embeddings
can be integrated with various anomaly detection models such as
one-class SVM \cite{chen2001one} and SVDD \cite{tax2004support}.  In
this work, to illustrate the effectiveness of learned low-dimensional
patch embeddings, we adopt the nearest neighbor rule in the embedding
space to localize anomalous pixels. This was also used in
\cite{cohen2020sub, yi2020patch}.

For every patch $p$ with stride
$S$ in test image $X$, we use its $L_2$ distance to the nearest normal
patch in the embedding space to define its anomaly score:
\begin{equation}\label{eq11}
AS (\bm{p}) = \min\ L_2(\bm{p}, \bm{p}_{Normal}).
\end{equation}
The pixel-wise anomaly score can be calculated by averaging anomaly
scores of all patches to which this pixel belongs to. An approximate
algorithm is adopted to mitigate the computational cost of the nearest
neighbor search. The maximum anomaly score of pixels in an image is set
to the image-level anomaly score. 

\begin{table*}[!htbp]
\begin{center}
\caption{Comparison of image anomaly localization
performance, where the evaluation metric is pixel-wise AUC-ROC. The best
and the second best results are shown in bold face and with an
underline, respectively.}\label{tab1}
\begin{adjustbox}{width=1.0\textwidth,center}
    \begin{tabular}{|l|c|c|c|c|c|c|c|c|c|c}
      \hline
      \textbf{} & AE L2 & AE SSIM & AnoGAN & VAE & SPADE & Patch-SVDD & FCDD &  PEDENet (ours) \\
      \hline
      Bottle & 0.86 & 0.93& 0.86 & 0.831 &  \textbf{0.984} &  \underline{0.981} & 0.80 & \textbf{0.984} \\
      Cable & 0.86 & 0.82& 0.78 & 0.831 &  \textbf{0.972} &   0.968 & 0.80 & \underline{0.971} \\
      Capsule & 0.88 & 0.94& 0.84 & 0.817 & \textbf{0.990}  & \underline{0.958} & 0.88 & 0.943 \\
      Hazelnut & 0.95 & 0.97& 0.87 & 0.877 & \textbf{0.991}   & \underline{0.975} & 0.96 & 0.970 \\
      Metal & 0.86 & 0.89& 0.76 & 0.787 & \textbf{0.981} &   \underline{0.980} & 0.88 & 0.973 \\
      Pill & 0.85 & 0.91& 0.87 & 0.813 & \textbf{0.965} &    0.951 & 0.86 & \underline{0.960} \\
      Screw & 0.96 & 0.96& 0.80 & 0.753 & \textbf{0.989} &   0.957 & 0.87 & \underline{0.972} \\
      Teeth brush & 0.93 &0.92 & 0.90 & 0.919 & \underline{0.979} & \textbf{0.981} & 0.90 & \underline{0.979}\\
      Transistor & 0.86 & 0.90& 0.80 & 0.754 & 0.941& \underline{0.970} & 0.80 & \textbf{0.982} \\
      Zipper & 0.77 & 0.88& 0.78 & 0.716 & \textbf{0.965} & 0.951 & 0.81 & \underline{0.962} \\ \hline
      \textbf{All 10 Object Classes} & 0.88 & 0.91 & 0.83  & 0.810 & \textbf{0.976} &  0.967 & 0.86 & \underline{0.970} \\ \hline
      Carpet & 0.59 & 0.87 & 0.54 & 0.597 & \textbf{0.975} & 0.926 & \underline{0.93} & 0.922 \\
      Grid & 0.90 & 0.94 & 0.58 & 0.612 & 0.937 & \textbf{0.962} & 0.87 & \underline{0.959} \\ 
      Leather & 0.75 & 0.78 & 0.64 & 0.671 & \underline{0.976} & 0.974 & \textbf{0.98} & \underline{0.976} \\
      Tile & 0.51 & 0.59 & 0.50 & 0.513 & 0.874 & 0.914 & \underline{0.92} & \textbf{0.926} \\
      Wood & 0.73 & 0.73 & 0.62 & 0.666 & 0.885 & \textbf{0.908} & 0.89 & \underline{0.900} \\ \hline
      \textbf{All 5 Texture Classes} & 0.70 & 0.78 & 0.29 & 0.612 & 0.929 & \textbf{0.937}  & 0.92 & \underline{0.936}\\ \hline
      \textbf{Average of 15 Classes} & 0.82 & 0.87 & 0.74 &  0.744 & \textbf{0.965} & 0.957 & 0.88 & \underline{0.959} \\ \hline
    \end{tabular}
  \end{adjustbox}
  \end{center}
\end{table*}

\begin{table*}[!htbp]
\begin{center}
\caption{Comparison of image anomaly detection performance, where the
evaluation metric is the image-level AUC-ROC. The best and the second best results
are shown in bold face and with an underline, respectively.}\label{tab2}
    \begin{tabular}{|l|c|c|c|c|c|c|c|c|c|c}
      \hline
      \textbf{} & GANomaly & ITAE & Patch-SVDD & SPADE & MahalanobisAD  & PEDENet (ours) \\ \hline
      \textbf{Image-Level AUC-ROC} & 0.762 & 0.839 & 0.921 & 0.855 & \textbf{0.958} & \underline{0.928} \\
      \hline
    \end{tabular}
  \end{center}
\end{table*}

\begin{figure*}[!t]
\centering
\includegraphics[width=1.0\linewidth]{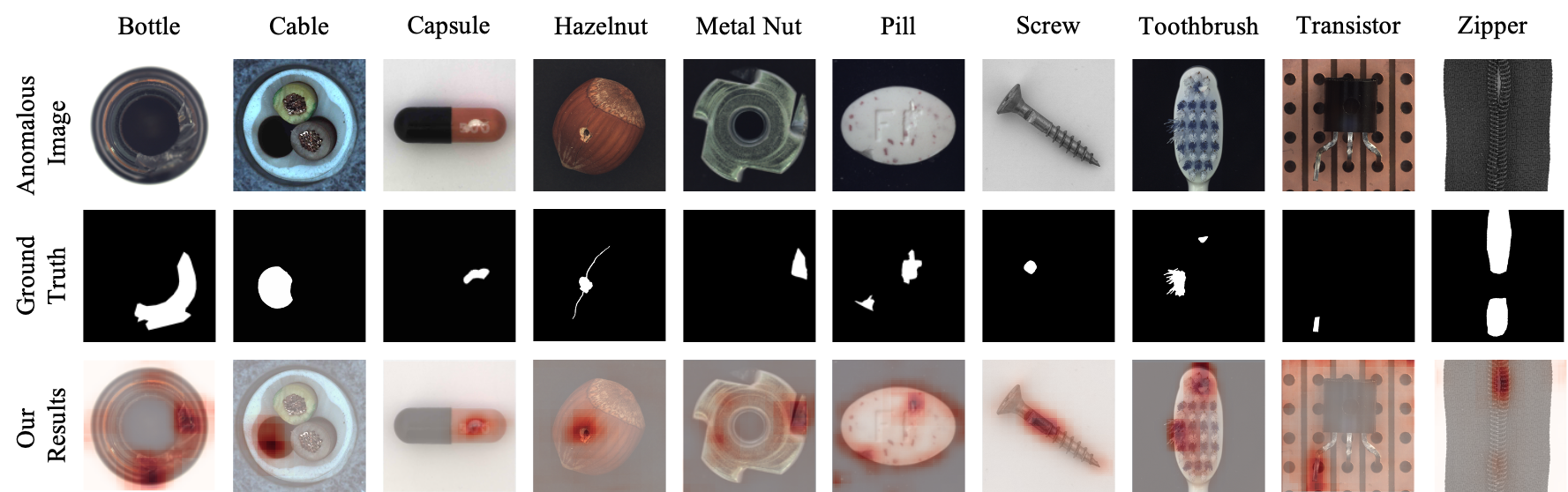}
\caption{Visualization of anomalous images, labeled ground truths and
localization results of the proposed PEDENet for 10 object classes in 
the MVTec AD dataset, where the red color is used to indicate detected 
anomaly regions.} \label{fig:5}
\end{figure*}

\begin{figure}[!t]
\centering
\includegraphics[width=9cm]{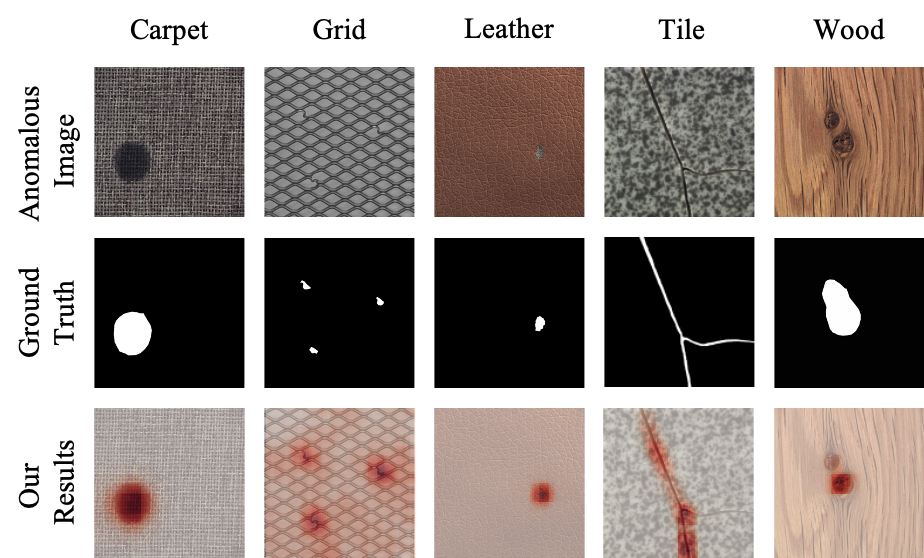}
\caption{Visualization of anomalous images, labeled ground truths and
localization results of the proposed PEDENet for 5 texture classes in 
the MVTec AD dataset, where the red color is used to indicate detected 
anomaly regions.} \label{fig:6}
\end{figure}

\section{Experiments}\label{sec:experiments}

\subsection{Experimental Setup}\label{subsec:setup}
To verify the effectiveness of the proposed PEDENet, we conducted
experiments on the MVTec AD dataset \cite{bergmann2019mvtec}, which is
a comprehensive anomaly localization dataset collected from real world
scenarios. It consists of images belonging to 15 classes, including 10
object classes and 5 texture classes. For each class, there are 60-391
training images and 40-167 test images, where image sizes varies from
$700\times700$ to $1024\times1024$. The training set contains only
normal images while the test set contains both normal and anomalous
images. Examples of normal and anomalous images are shown in Fig.
\ref{fig:1}. We train and evaluate anomaly localization algorithms for
each class separately, which is known as class-specific evaluation.  To
train a model, all training images are first resized to $256\times256$
and, then, patches of size $64\times64$ are randomly sampled from these
resized images. 

Our proposed solution consists of three networks: 1) the PE network, 2)
the DE network, and 3) the LP network. The PE network consists of eight
convolutional layers and one output layer. The filter size in all
convolutional layers is the same, i.e. $3\times3$. The filter numbers
for convolutional layers are 32, 64, 128, 128, 64, 64, 32, 32, 64 and
LeakyReLU \cite{he2015delving} with slope 0.1 is used as activation
function.  The {\em tanh} activation is used in the output layer to
normalize the output to the range of [-1.0,1.0]. Both the DE and LP
networks are multilayer perceptrons (MLPs) with the LeakyReLU activation
of slope 0.1.  The DE network has three hidden layers of 128, 64, 32
neurons, respectively. The LP network has two hidden layers of 128
neurons per layer. The input to the LP network is subtraction of
features from two neighboring patches. 

We train all networks using the Adam optimizer with learning rate
0.0001. The batch sizes are 128 image patches for the DE network and 36
pairs of adjacent patches for the LP network. All experiments are
conducted on a machine equipped with an Intel i7-5930K CPU and an NVIDIA
GeForce Titan X GPU. 


\subsection{Performance Evaluation}\label{subsec:performance}

\textbf{Anomaly Localization.} We compare the image anomaly localization
performance of several methods with respect to the MVTec AD dataset
\cite{bergmann2019mvtec} in Table \ref{tab1}, where the evaluation
metric is the pixel-wise area under the receiver operating
characteristic curve (AUC-ROC). The benchmarking methods are listed
below. 
\begin{itemize}
\item Reconstruction approach: AE L2 \cite{bergmann2019mvtec}, AE SSIM
\cite{DBLP:conf/visapp/BergmannLFSS19}, Variational Autoencoder(VAE) and AnoGAN \cite{schlegl2017unsupervised}. 
\item Pre-trained network based approach: SPADE \cite{cohen2020sub}. 
\item One-class classification approach: Patch-SVDD\cite{yi2020patch}, and
FCDD \cite{liznerski2020explainable}. 
\end{itemize}
Our proposed PEDENet achieves 97.0\% for the average of 10 object classes
(the 2nd best), 93.6\% for the average of 5 texture classes (the 2nd
best), and 95.9\% for the average of all 15 classes (the 2nd best).
Note that there is no single method that performs the best in all cases.
PEDENet is second to SPADE in the average of 10 object classes while it
is second to Patch-SVDD in the average of the 5 texture classes. The
performance differences among SPADE, Patch-SVDD and PEDENet are actually
quite small. Thus, it is fair to say that PEDENet is one of the
state-of-the-art methods for image anomaly localization. 

\textbf{Anomaly Detection.} We compare image anomaly detection
performance in Table \ref{tab2}, where the evaluation metric is the
image-level area under the receiver operating characteristic curve
(AUC-ROC).  The benchmarking methods include: GANomaly
\cite{akcay2018ganomaly}, ITAE \cite{fei2020attribute}, Patch-SVDD
\cite{yi2020patch}, SPADE \cite{cohen2020sub} and MahalanobisAD
\cite{rippel2020modeling}.  Our proposed PEDENnet reaches 92.8\%. It
outperforms both SPADE and Patch-SVDD.  It is only second to
MahalanobisAD. Thus, PEDENnet also offers state-of-the-art image anomaly
detection performance. 

\textbf{Visualization of Localized Anomalies.} Anomalous images, labeled
ground truths and localization results of the proposed PEDENet for 10
object classes and 5 texture classes in the MVTec AD dataset are
visualized in Figs.  \ref{fig:5} and \ref{fig:6}, respectively.
Anomalous regions are highlighted in red. A region is more likely to be
anomalous if it has stronger red color. As shown in those figures,
anomalous regions could be accurately detected and localized by the
proposed PEDENet. Note that relatively small and inconspicuous defects
can still be spotted such as Capsule, Hazelnut and Pill examples in Fig.
\ref{fig:5}. 

\textbf{Model Size.} Most state-of-the-art methods that give the best
performance take the pre-trained network approach such as SPADE
\cite{cohen2020sub}, and MahalanobisAD \cite{rippel2020modeling}.  As a
result, their model sizes are usually quite large. We compare the model
sizes of these three methods in Table \ref{tab3} in terms of the number
of model parameters. The numbers of model parameters of SPADE and
MahalanobisAD are 145$\times$ and 36$\times$ of that of PEDENet.  A
smaller model size means lower memory requirement which is critical in
real-world deployment. Besides being a light-weighted network, PEDENet
is trained using the training data in the MVTec AD dataset only. In
contrast, the two benchmarking methods leverage giant models that are
pre-trained by millions of labeled images. 

\begin{table}[htbp]
  \begin{center}
    \caption{Model Size Comparison}\label{tab3} 
    \begin{tabular}{|l|c|c|} \hline
      \textbf{Methods} & Pre-trained Model & \# of Parameters \\ \hline
      SPADE & Wide ResNet-50 & 68M \\ \hline
      MahalanobisAD & EfficientNet-B4 & 17M \\ \hline
      PEDENet (ours) & None & 0.47M \\ \hline
    \end{tabular}
  \end{center}
\end{table}
\begin{table}[htb]
  \begin{center}
    \caption{Ablation Study}\label{tab4} 
    \begin{tabular}{c|c|c} \hline
      $\mathcal{L}_{DEN}$ & $\mathcal{L}_{LPN}$ & AUC-ROC \\ \hline
      $\surd$ & & 0.810 \\
      & $\surd$ & 0.894 \\
      $\surd$ & $\surd$ & 0.959 \\ \hline
    \end{tabular}
  \end{center}
\end{table}
\textbf{Ablation Study.} We conduct ablation study to understand the
impact of each loss term in Eq. (\ref{eq10}) to the performance of the
proposed PEDENet. That is, we remove either $\mathcal{L}_{DEN}$ or
$\mathcal{L}_{LPN}$ and train the model under the same condition. We see
from Table \ref{tab4} that the adoption of both $\mathcal{L}_{DEN}$ and
$\mathcal{L}_{DEN}$ as shown in Eq. (\ref{eq10}) improves the anomaly
localization performance.

\textbf{Challenge of Texture Classes.} It is worthwhile to point out
that almost all methods have an obvious performance gap between the
object classes and the texture classes. As shown in Table \ref{tab1},
the performance gap varies from $3\%$ \cite{yi2020patch} to almost
$20\%$ \cite{DBLP:conf/visapp/BergmannLFSS19}.  As mentioned in
\cite{yi2020patch}, the optimal hyper-parameters for objects and
textures are likely to be different and there is a trade-off to get the
best average performance.  Actually, texture and regular images are
treated differently in image processing. There many unique methods
developed specifically for textures \cite{chang1993texture,
zhang2019texture, zhang2019data, zhang2021dynamic}.  Unlike regular
images, there are strong self-similarity and quasi-periodicity in
textures, which could be exploited to achieve further performance
improvement. 

\section{Conclusion and Future Work}\label{sec:conclusion}

A new neural network model, called the PEDENet, was proposed for image
anomaly detection and localization. It is a lightweight model and offers
state-of-the-art anomaly detection and localization performance. There
are some future research topics. First, the MVTec AD dataset is still
too small. It is valuable to collect more samples and build a larger
dataset.  Second, it is desired to find a more effective solution for
texture anomaly detection and localization. Third, there are many
hyper-parameters in the proposed PEDENnet, which are finetuned by trial
and error. A more systematic approach is needed. Finally, it is
interesting to find an altenative image anomaly detection and
localization solution that is effective and interpretable.

\bibliographystyle{unsrt}  
\bibliography{references}

\end{document}